\documentclass[11pt,a4paper]{article}
\usepackage[hyperref]{acl2017}
\usepackage{url}
\usepackage{times}
\usepackage{latexsym}
\usepackage{amsmath}
\usepackage{float}
\usepackage{graphicx}
\usepackage{framed}

\restylefloat{figure}

\aclfinalcopy
\def\confidential{DRAFT COPY.  DO NOT DISTRIBUTE.}

\title{SLING: A framework for frame semantic parsing}

\author{
Michael Ringgaard \\ Google Inc. \\ {\tt ringgaard@google.com} \\\And
Rahul Gupta \\ Google Inc. \\ {\tt grahul@google.com} \\\And
Fernando C. N. Pereira \\ Google Inc. \\ {\tt pereira@google.com} \\
}

\begin{document}
\maketitle

\begin{abstract}
We describe SLING, a framework for parsing natural language into
semantic frames. SLING supports general transition-based, neural-network parsing
with bidirectional LSTM input encoding and a Transition Based Recurrent
Unit (TBRU) for output decoding. The parsing model is
trained end-to-end using only the text tokens as input. The
transition system has been designed to output frame graphs directly without
any intervening symbolic representation.
The SLING framework includes an efficient and scalable frame store
implementation as well as a neural network JIT compiler for fast inference
during parsing.
SLING is implemented in C++ and it is available for download on GitHub.
\end{abstract}

\section{Introduction}

Recent advances in machine learning make it practical to train
recurrent multi-level neural network classifiers, allowing us to rethink the
design and implementation of natural language
understanding (NLU) systems.

Earlier machine-learned NLU systems were commonly organized as pipelines of
separately trained stages for syntactic and semantic annotation of text.
A typical  pipeline would start with part-of-speech (POS) tagging, followed by
constituency or dependency parsing for syntactic analysis.
Using the POS tags and parse trees as feature inputs, later stages in the
pipeline could then derive semantically relevant annotations such as entity and
concept mentions, entity types, coreference relationships, and semantic roles
(SRL).

For simplicity and efficiency, each stage in a practical NLU pipeline would just
output its best hypothesis and pass it on to the next stage~\cite{finkel2006}.
Obviously, errors could then accumulate
throughout the pipeline making it much harder for the system to perform
accurately. For instance, F1 on SRL drops by more than 10\% when going from gold
to system parse trees~\cite{toutanova2005}.

However, applications may not need the intermediate annotations produced
by the earlier stages of a NLU pipeline, so it would be preferable if all stages
could be trained together to optimize an objective based on the output
annotations needed for a particular application.

Earlier NLU pipelines often used linear classifiers for each stage.
Linear classifiers achieve simplicity and training efficiency at the expense of
feature complexity, requiring elaborate feature
extraction, many different feature types, and
feature combinations to achieve reasonable accuracy.
With deep learning, we can use embeddings, multiple layers, and recurrent
network connections to reduce the need for complex
feature design. The internal learned representations in model hidden layers
replace the hand-crafted feature combinations and intermediate representations
in pipelined systems.

The SLING parser exploits deep learning to bypass those limitations of classic
pipelined systems.
It is a transition-based parser that outputs frame graphs directly without any
intervening symbolic representation (see Section~\ref{sec:ts}). Transition-based
parsing is often associated with dependency parsing, but we have designed a
specialized transition system that outputs frame graphs instead of dependency
trees.

\begin{figure*}[t]
  \centering
  \includegraphics[width=1.0\linewidth]{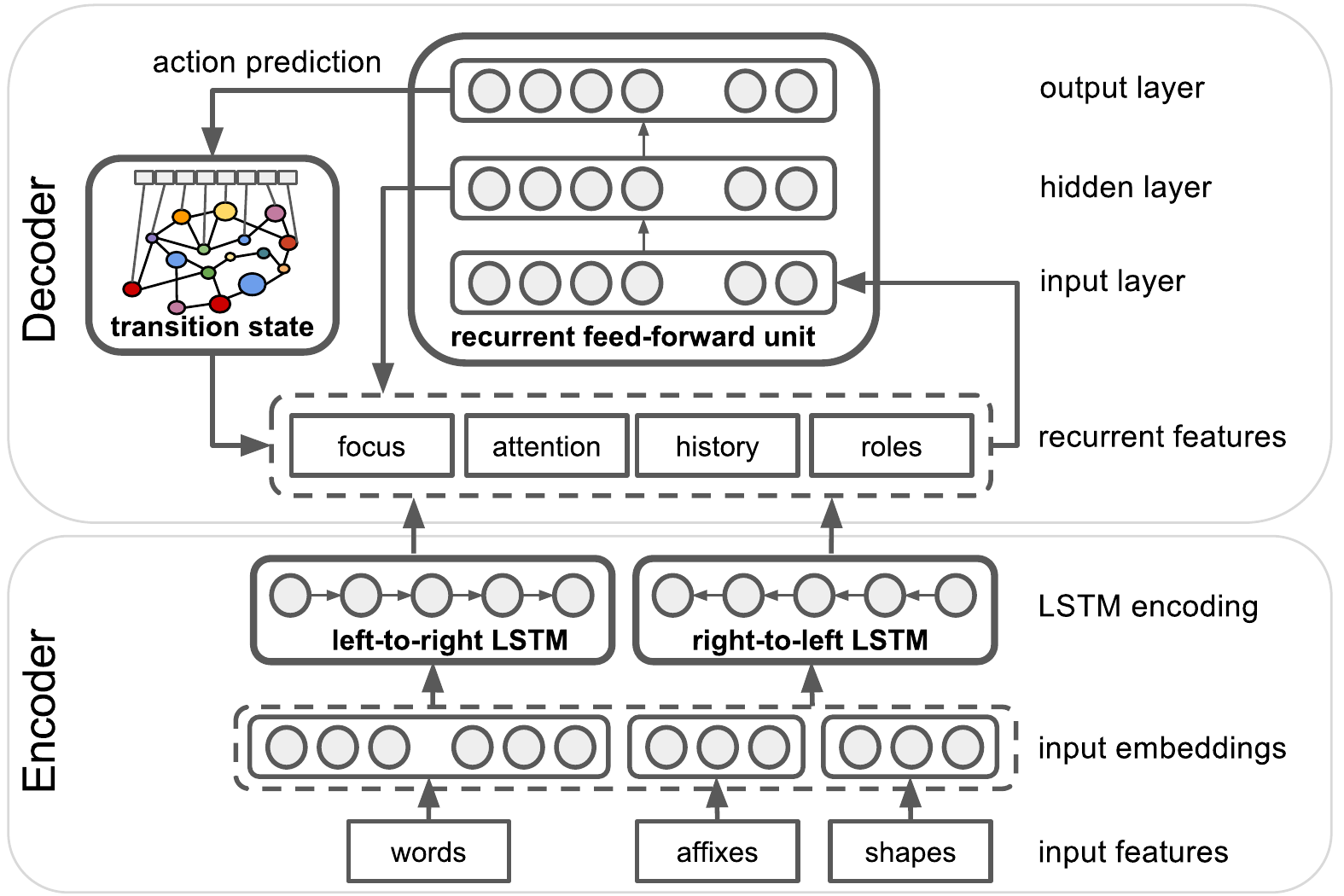}
  \caption{Neural network architecture of the SLING parser. The input is encoded
  by a bi-directional LSTM and fed into a recurrent feed-forward (FF) unit that
  proposes transition system actions.
  The hidden layer activations and the transition system state are combined to
  create the input feature vector for the next step. The FF unit is
  run repeatedly until the transition system has reached a final state.}
  \label{fig:network}
\end{figure*}

We use a recurrent feed-forward unit for predicting the actions in the
transition sequence, where the hidden activations from predicting each
transition step are fed back into subsequent steps.
A bidirectional LSTM (biLSTM) encodes the input into a sequence of vectors(Figure~\ref{fig:network}).
This neural network architecture has been implemented using DRAGNN~\cite{dragnn}
and TensorFlow~\cite{tensorflow}.

The SLING framework and a semantic parser built in it are now available as
open-source code on GitHub.\footnote{\url{https://github.com/google/sling}}

In Section~\ref{sec:framesem} we introduce \emph{frame semantics}, the
linguistic theory that inspired SLING, as well as the SLING frame store, a
C++ framework for representing and storing semantic frames compactly and
efficiently.
Section~\ref{sec:att} introduces the parser's frame-semantics-oriented attention
mechanism, and Section~\ref{sec:ts} describes the transition system used for
producing frame graphs. Section~\ref{sec:features} describes the features used
by the parser. In sections~\ref{sec:experiments} and~\ref{sec:eval} we describe
our experiments on OntoNotes, and section~\ref{sec:runtime} describes the
fast parser runtime.

\section{Frame semantics}
\label{sec:framesem}

While frames in SLING are not tied to any particular linguistic theory or
knowledge ontology, they are inspired by \emph{frame semantics}, the
theory of linguistic meaning originally developed by Charles Fillmore~\cite{fillmore1982}.
Frame semantics connects linguistic semantics to encyclopedic knowledge, with the
central idea that understanding the meaning of a word requires access to all
the essential knowledge that relates to that word. A word \emph{evokes} a frame
representing the specific concept it refers to.

A semantic frame is a set of statements that give "characteristic
features, attributes, and functions of a denotatum, and its characteristic
interactions with things necessarily or typically associated with it."~\cite{alan2001}.
A semantic frame can also be viewed as a coherent group of concepts
such that complete knowledge of one of them requires knowledge of all of them.

Frame semantics is not just for individual concepts, but can be generalized
to phrases, entities, constructions, and other larger and more complex linguistic
and ontological units. Semantic frames can also model world knowledge and inferential relationships
in common sense,
metaphor~\cite{narayanan1999},
metonymy,
action~\cite{narayanan1999reasoning},
and perspective~\cite{chang2002}.

\section{Frames in SLING}
\label{sec:slingframes}

SLING represents frames with data structures consisting of a list of slots, where each
slot has a name (role) and a value. The slot values can be literals like numbers
and strings, or links to other frames. A collection of interlinked frames can thus be seen as a directed
graph where the frames are the (typed) nodes and the slots are the (labeled)
edges. A frame graph can also be viewed as a feature structure~\cite{carpenter2005}
and unification can be used for induction of new frames from existing frames.
Frames can also be used to represent more basic data
structures such as a C struct with fields, a JSON object, or a record in a
database.

SLING frames live inside a \emph{frame store}. A store is a container that
tracks all the frames that have been allocated in the store, and serves as a
memory allocation arena for them. When making a new frame, one
specifies the store where the frame should be allocated. The frame will live in
this store until the store is deleted or the frame is garbage collected because
there no remaining live references to it.\footnote{See the \href{https://github.com/google/sling/blob/master/frame/README.md}{SLING Guide}
for a detailed description of the SLING frame store implementation.}

SLING frames are externally represented in a superset of JSON that allows
references between frames (JSON objects) with the {\tt \#n} syntax. Frames can
be assigned identifiers (\emph{ids}) using the {\tt=\#n} syntax. SLING frames
can  have both numeric and named ids and both slot names and values can be frame
references. Where JSON objects can only represent trees, SLING frames can be
used for representing arbitrary graphs. SLING has special syntax for built-in
slot names:

\begin{table}[h!]
\begin{tabular}{|l|l|l|}
\hline
Syntax & Symbol & RDF \\
\hline
=name & id:name & rdf:ID \\
:name & isa:name & rdf:InstanceOf \\
+name & is:name & rdfs:subClassOf \\
\hline
\end{tabular}
\end{table}

Documents are also represented using frames, where the document frame has slots
for the document text, the tokens, and the mention phrases and the frames they
evoke. See Figure~\ref{fig:slingdoc} for an example.

\begin{figure*}[t]
\begin{framed}
\begin{verbatim}
{
  :/s/document
  /s/document/text: "John hit the ball"
  /s/document/tokens: [
    {/s/token/text: "John" /s/token/start: 0  /s/token/length: 4},
    {/s/token/text: "hit"  /s/token/start: 5  /s/token/length: 3},
    {/s/token/text: "the"  /s/token/start: 9  /s/token/length: 3},
    {/s/token/text: "ball" /s/token/start: 13 /s/token/length: 4}
  ]
  /s/document/mention: {
    :/s/phrase /s/phrase/begin: 0
    /s/phrase/evokes: {=#1 :/saft/person }
  }
  /s/document/mention: {
    :/s/phrase /s/phrase/begin: 1
    /s/phrase/evokes: {
      :/pb/hit-01
      /pb/arg0: #1
      /pb/arg1: #2
    }
  }
  /s/document/mention: {
    :/s/phrase /s/phrase/begin: 3
    /s/phrase/evokes: {=#2 :/saft/consumer_good }
  }
}
\end{verbatim}
\end{framed}
\caption{The text ``John hit the ball" in SLING frame notation. The document
itself is represented by a frame that has the text, an array of tokens and
the mentions that evoke frames. There are three frames: a person frame (John),
a consumer good frame (bat) and a hit-01 frame. The hit frame has the person
frame as the agent (arg0) and the ball frame as the object (arg1).}
\label{fig:slingdoc}
\end{figure*}

\section{Attention}
\label{sec:att}

The SLING parser is a kind of sequence-to-sequence model that first encodes the
input text token sequence with a bidirectional LSTM encoder and then runs
the transition system on that encoding to produce a sequence of transitions,
where each transition updates the system state that combined with the input
encoding form the input for the transition feed-forward cell that predicts the
next transition (Figure~\ref{fig:network}).

Sequence-to-sequence models often rely on an ``attention" mechanism to focus
the decoder on the parts of the input most relevant for producing the next
output symbol. In this work, however, we use a somewhat difference attention
mechanism, loosely inspired on neuroscience models of attention and awareness
~\cite{nelson2017,graziano2013}. In our model, attention focuses on parts of the
frame representation that the parser has created so far, rather than focusing
on (encodings of) input tokens as is common for other sequence-to-sequence
attention mechanisms.

We maintain an \emph{attention buffer} as part of the transition system state.
This an ordered list of frames, where the order represents closeness to the
center of attention. Transition system actions maintain the attention buffer, bringing
a frame to the front when the frame is evoked or re-evoked by the input text.
When a new frame is evoked, it will merge the concept and its roles into a new
coherent chunk of meaning, which is represented by the new frame and its
relations to other frames, and this will become the new center of attention.
Our hypothesis is that by maintaining this attention mechanism, we only need to
look at a few recent frames brought into attention to build the desired
frame graph.

\section{Transition system}
\label{sec:ts}

\emph{Transition systems} are widely
used in parsing to build dependency parse trees as a side effect of performing a sequence \emph{state transitions}
$(s_i,a_i)$ where $s_i$ is a \emph{state} and $a_i$ is an \emph{action}. Action $a_i$ computes the new state
$s_{i+1}$ from state $s_i$. For example, the \emph{arc-standard}
transition system~\cite{nivre2006} uses a sequence of {\bf SHIFT}, {\bf LEFT-ARC(label)}, and
{\bf RIGHT-ARC(label)} actions, operating on a state whose main component is a stack, to build a dependency parse tree.

We use the same idea to construct a frame graph where frames can be
evoked by phrases in the input. But instead of using a stack in the state, we use the  attention
buffer introduced in the previous section that keeps track of the most salient
frames in the discourse.

The attention buffer is a priority list of all the
frames evoked so far. The front of the buffer serves as the working
memory for the parser. Actions operate on the front of the buffer and in some cases other frames in the buffer. The transition
system simultaneously builds the frame graph and maintains the attention buffer
by moving the frame involved involved in an action to the front of the attention
buffer. At any time, each evoked frame has a unique position in the attention
buffer.

The transition system consists of the following actions:

\begin{itemize}
  \item {\bf SHIFT} -- Moves to next input token. Only valid when not at the
        end of the input buffer.
  \item {\bf STOP} -- Signals that we have reach the end of the parse. This is
        only valid when at the end of the input buffer. Multiple STOP actions
        can be added to the transition sequence, e.g. to make all sequences in a
        beam have the same length. After a STOP is issued, no other actions are
        permitted except more STOP actions.
  \item {\bf EVOKE(type, n)} -- Evokes a frame of type {\bf type} from
        the next {\bf n} tokens in the input. The evoked frame is inserted at the front of the attention
        buffer, becoming the new center of attention.
  \item {\bf REFER(frame, n)} -- Makes a new mention from the next {\bf n} tokens
        in the input evoking an existing frame in the attention buffer. This
        frame is moved to the front of the attention buffer and will become the
        new center of attention.
  \item {\bf CONNECT(source, role, target)} -- Adds slot to {\bf source} frame
        in the attention buffer with name {\bf role} and value {\bf target}
        where {\bf target} is an existing frame in the attention buffer. The
        {\bf source} frame become the new center of attention.
  \item {\bf ASSIGN(source, role, value)} -- Adds slot to {\bf source} frame in
        the attention buffer with name {\bf role} and constant value {\bf value}
        and moves the frame to the front of the buffer. This action
        is only used for assigning a constant value to a slot, in contrast to
        {\bf CONNECT} where the value is another frame in the attention buffer.
  \item {\bf EMBED(target, role, type)} -- Creates a new frame with
        type {\bf type} and adds a slot to it with name {\bf role} and value
        {\bf target} where {\bf target} is an existing frame in the attention
        buffer. The new frame becomes the center of attention.
  \item {\bf ELABORATE(source, role, type)} -- Creates a new frame with type
        {\bf type} and adds a slot to an existing frame {\bf source} in the
        attention buffer with {\bf role} set to the new frame. The new frame
        becomes the center of attention.
\end{itemize}

In summary, {\bf EVOKE} and {\bf REFER} are used to evoke frames from text
mentions, while {\bf ELABORATE} and {\bf EMBED} are used to create frames not
directly evoked by text.

This transition system can generate any connected frame graph where the frames
are either directly on indirectly evoked by phrases in the text. A frame
can be evoked by multiple mentions and the graph can have cycles.

The transition system can potentially have an unbounded number of actions since
it is parameterized by phrase length and attention buffer indices which can be
arbitrarily large. In the current implementation, we only consider the
top $k$ frames in the attention buffer ($k=5$) and we do not consider any phrases
longer than those in the training corpus.

Multiple transition sequences can generate the same frame annotations, but we
have implemented an oracle sequence generator that takes a document and converts
it to a canonical transition sequence in a way similar to how this is done
for transition-based dependency parsing~\cite{nivre2006}. For example, the sentence
``John hit the ball" generates the following transition sequence:
\begin{verbatim}
  EVOKE(/saft/person, 1)
  SHIFT
  EVOKE(/pb/hit-01, 1)
  CONNECT(0, /pb/arg0, 1)
  SHIFT
  SHIFT
  EVOKE(/saft/consumer_good, 1)
  CONNECT(1, /pb/arg1, 0)
  SHIFT
  STOP
\end{verbatim}

\section{Features}
\label{sec:features}

The biLSTM uses only lexical features based on the current input word:

\begin{itemize}
  \item The current word itself. During training we initialize the embedding
  for this feature from pre-trained word embeddings~\cite{mikolov2013} for all
  the words in the the training data.
  \item The prefixes and suffixes of the current input word. We use only
  prefixes up to three characters in our experiments.
  \item Word shape features based on the characters in the current input word:
  hyphenation, capitalization, punctuation, quotes, and digits. Each of these
  features has its own embedding matrix.
\end{itemize}

The TBRU is a simple feed-forward unit with a single hidden layer.
It takes the hidden activations from the biLSTM as well as the activations from
the hidden layer from the previous steps as raw input features, and maps them
through embedding matrices to get the input vector for the  hidden layer. More specifically,
the inputs to the TBRU are as follows:

\begin{itemize}
  \item The left-to-right and right-to-left LSTMs supply their activations
  for the current token in the parser state.
  \item The attention feature looks at the top-$k$ frames in the attention buffer
  and finds the phrases in the text (if any) that evoked them. The activations
  from the left-to-right and right-to-left LSTMs for the last token of each of those
  phrases are are included as TBRU inputs, serving as continuous lexical
  representations of the top-$k$ frames in the attention buffer.
  \item The hidden layer activations of the transition steps which evoked or
  brought into focus the top-$k$ frames in the attention buffer are also inputs to the TBRU,
  providing a
  continuous representation for the semantic frame contexts that evoked those frames most recently.
  \item The history feature uses the hidden activations in the feed-forward
  unit from the previous $k$ steps as feature inputs to the current step.
  \item Embeddings of triples of the form $(s_i, r_i, t_i)$, $0<s_i,t_i\le k$ encode the fact that the frame at position $s_i$ in the attention buffer has a role $r_i$ with
  the frame at position $t_i$ in the attention buffer as its value. Back-off
  features are added for the source roles $(s_i,r_i)$, target role $(r_i, t_i)$,
  and unlabeled roles $(s_i,t_i)$.
\end{itemize}

\section{Experiments}
\label{sec:experiments}

We derived a corpus annotated with semantic frames from the OntoNotes corpus~\cite{ontonotes2006}. We took the
PropBank SRL layer~\cite{palmer2005} and converted the predicate-argument
structures into frame annotations. We also annotated the corpus with
entity frames based on entity types from a state-of-the-art entity tagger.
We determined the head token of each argument span and if this coincided
with the span of an existing frame, then we used it as the evoking span for the
argument frame, otherwise we just used the head token as the evoking span of the
argument frame.

The various frame types mentioned above are listed in
Table~\ref{tab:types}. They include 7 conventional entity types,
6 top-level non-entity types (e.g. date), 13 measurement types, and
more than 5400 PropBank frame types. All the frame roles are collapsed onto
/pb/arg0, /pb/arg1, and so on. Our training corpus size was $111,006$
sentences, $2,206,274$ tokens.

\begin{table*}[t]
\begin{tabular}{|l|p{11cm}|}
\hline
{\bf Type set} & {\bf Details} \\
\hline
Entity types & /saft/\{person, location, organization, art, consumer\_good, event, other\} \\
\hline
Top-level non-entity types & /s/\{thing, date, price, measure, time, number\} \\
\hline
Fine-grained measure types & /s/measure/\{area, data, duration, energy, frequency, fuel, length, mass, power, speed, temperate, voltage, volume\} \\
\hline
PropBank SRL types & 5426 types, e.g. /pb/write-01, /pb/tune-02  \\
\hline
\end{tabular}
\caption{Frame types used in the experiments.}
\label{tab:types}
\end{table*}

\begin{table}[t]
\begin{tabular}{|l|r|r|}
\hline
{\bf Action Type} & {\bf \# Unique Args} & {\bf Raw Count} \\
\hline
SHIFT & 1 & 2,206,274 \\
\hline
STOP & 1 & 111,006 \\
\hline
EVOKE & 5,532 & 1,080,365 \\
\hline
CONNECT & 1,421 & 635,734 \\
\hline
ASSIGN & 13 & 5,430 \\
\hline
{\bf Total} & 6,968 & 4,038,809 \\
\hline
\end{tabular}
\caption{Action statistics for the transitions that generated
the gold frames for the OntoNotes training corpus.}
\label{tab:action-table}
\end{table}

Table~\ref{tab:action-table} shows action statistics for the transition sequences that
generate the gold frames in the training corpus. As expected, there is one
SHIFT action per training token, and one STOP action per training sentence.
The EVOKE action occurred with $5,532$ unique (length, type) arguments in the
corpus, for a raw count of roughly $1.08$ million action tokens. Overall our action space
had $6968$ action types, which is also the size of the softmax layer of our TBRU
decoder.

\noindent{{\bf Hyperparameters:}} Our final set of hyperparameters after
grid search with a dev corpus was: $\mbox{learning\_rate} = 0.0005$,
$\mbox{optimizer} = \mbox{Adam}$~\cite{kingma2014} with $\beta_1 = 0.01$, $\beta_2 = 0.999$,
$\epsilon = 1e-5$, no dropout, gradient clipping at $1.0$, exponential moving
average, no layer normalization, and a training batch size of $8$. We use
$32$ dimensional word embeddings, single layer LSTMs with $256$ dimensions,
and a $128$ dimensional hidden layer in the feed-forward unit.

\begin{figure}
\centering
\includegraphics[width=\columnwidth]{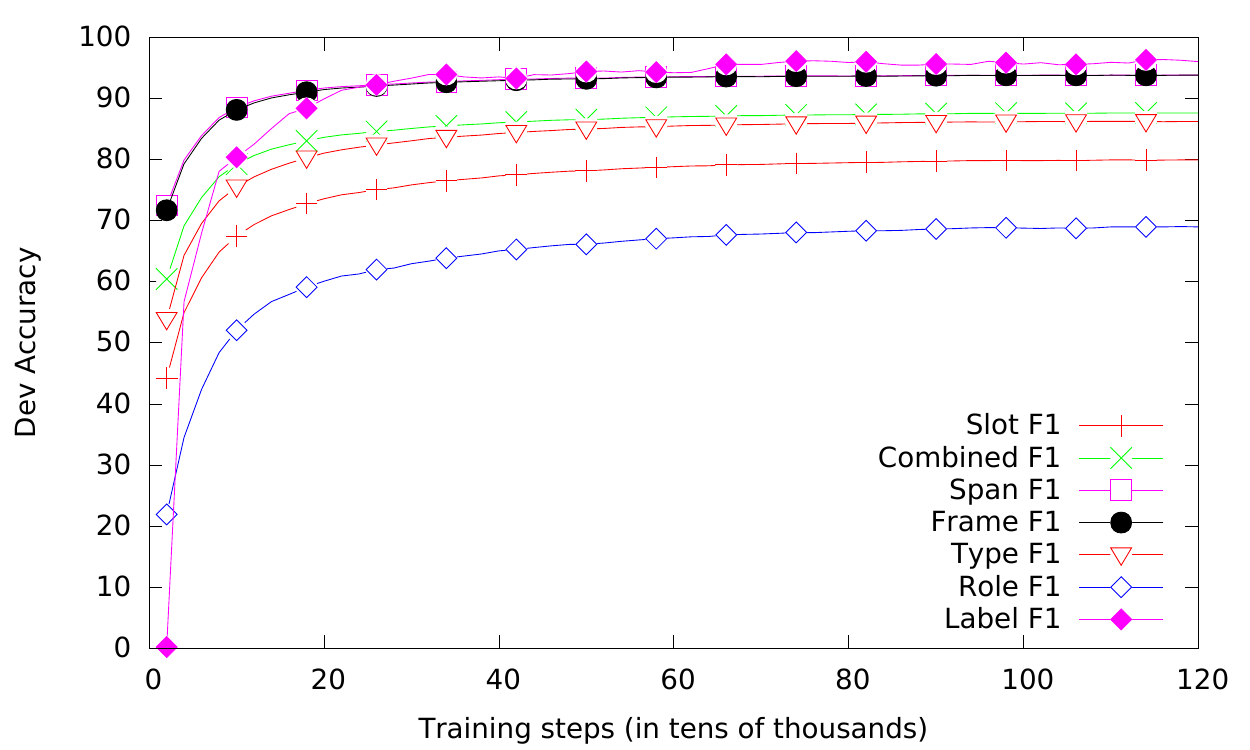}
  \caption{Frame graph quality metrics on dev set as training progresses.
	Training was stopped at $120,000$ iterations since we saw very little
	improvement after that.}
  \label{fig:dev-eval}
\end{figure}

We stopped training after $120,000$ steps, where each step corresponds to
processing one training batch, and evaluated on the dev corpus
($15,084$ sentences) after every checkpoint (= $2,000$ steps).
Figure~\ref{fig:dev-eval} shows the how the various evaluation metrics evolve
as training progresses. Section~\ref{sec:eval} contains the details of these
metrics are evaluated. We picked the checkpoint with the best `Slot F1` score.

\section{Evaluation}
\label{sec:eval}

An annotated document consists of a number of connected frames as well as
phrases (token spans) that evoked these frames. We evaluated
annotation quality by comparing the generated frames with the gold standard frame
annotations from the evaluation corpus.

Two documents are matched by constructing a virtual graph where the document
is the start node. The document node is then connected to the spans and the
spans are connected to the frames that the spans evoke. This graph is then
extended by following the frame-to-frame links via the roles. Quality is
computed by aligning the golden and predicted graphs and computing precision,
recall, and F1. Those scores are separately computed for spans, frames,
frame types, roles that link to other frames (referred to as 'roles'),
and roles that link to global constants (referred to as 'labels').

We also report two aggregate quality scores: (a) {\em Slot}, which is
an aggregate of {\em Type}, {\em Role}, and {\em Label}, and (b) {\em Combined},
which is an aggregate of {\em Span}, {\em Frame}, {\em Type}, {\em Role}, and
{\em Label}.

We rated the checkpoints using the Slot-F1 metric and selected the checkpoint with
the best Slot-F1. Intuitively, a high {\em Slot} score reflects that the
right type of frames are being evoked, along with the right set of slots and
links to other frames.

Figure~\ref{fig:dev-eval} shows that as training progresses,
the model learns to output the spans and frames evoked from those spans with
fairly good quality (SPAN F1 $\approx$ FRAME F1 $\approx$ $93.81\%$). It also
gets the type of those frames right with a TYPE F1 of $= 85.88\%$. ROLE F1
though is lower at just $69.65\%$. ROLE F1 measures the accuracy of correctly
getting the frame-frame link, including the label of the link. Further error
analysis will be required to understand how frame-frame links are missed by
the model. Also note that currently the {\em roles} feature is the only one
that captures inter-frame link information.
Augmenting this with more features should help improve ROLE quality, as we will
investigate in future work.

Finally, we took the best checkpoint, with SLOT F1 $= 79.95\%$ at $118,000$ steps,
and evaluated it on the test corpus.
Table~\ref{tab:eval} lists the quality of this model on the test and dev
corpora.
With the exception of LABEL accuracies, all the other metrics exhibit less than
half a percent difference between the test and dev corpora. This illustrates
that despite the lack of dropout, the model generalizes well to unseen text.
As for the disparity on LABEL F1 ($95.73$ on dev
against $92.81$ on test), we observe from Figure~\ref{fig:dev-eval}
that the LABEL accuracies follow a different improvement pattern
during training. On the dev set, LABEL F1 peaked at $96.18$ at $100,000$ steps,
and started degrading slightly from there on to $95.73$ at $118,000$ steps,
possibly showing signs of overfitting which are absent in the other metrics.

\begin{table}[ht]
\begin{tabular}{|ll|r|r|}
\hline
{\bf Metric} & & {\bf Dev} & {\bf Test} \\
\hline
Tokens & & 291,746  & 216,473  \\
\hline
Sentences &   & 15,084 & 11,623  \\
\hline
\hline
Span & Precision & 93.42  & 93.04 \\
\hline
& Recall & 94.21 & 94.34 \\
\hline
& F1 & 93.81 & 93.69 \\
\hline
Frame & Precision & 93.47 & 93.20 \\
\hline
& Recall & 94.16 & 94.08 \\
\hline
& F1 & 93.81 & 93.64\\
\hline
Type & Precision & 85.56  & 85.67 \\
\hline
& Recall & 86.20 & 86.49 \\
\hline
& F1 & 85.88 & 86.08 \\
\hline
Role & Precision & 70.21 & 69.59 \\
\hline
& Recall & 69.11 & 69.20 \\
\hline
& F1 & 69.65 & 69.39 \\
\hline
Label & Precision & 96.51 & 95.02 \\
\hline
& Recall & 94.97 & 90.70 \\
\hline
& F1 & 95.73 & 92.81 \\
\hline
Slot & Precision & 80.00 & 79.81 \\
\hline
& Recall & 79.90 & 80.10 \\
\hline
& F1 & 79.95 & 79.96 \\
\hline
Combined & Precision & 87.46 & 87.20 \\
\hline
& Recall & 87.79 & 87.91 \\
\hline
& F1 & 87.63 & 87.55 \\
\hline
\end{tabular}
\caption{Evaluation on dev and test corpora, model
chosen on the Slot-F1 metric on dev corpus.}
\label{tab:eval}
\end{table}

We have tried increasing the sizes of the LSTM dimensions, hidden layers,
and embeddings, but this did not improve the results significantly.

\section{Parser runtime}
\label{sec:runtime}

The SLING parser uses TensorFlow~\cite{tensorflow} for training but it also
supports annotating text with frame annotations at runtime. It can take
advantage of batching and multi-threading to speed up parsing. However, in
practical applications of the parser, it may not be convenient to
batch documents for processing, so to have a realistic benchmark, we
set the batch size to one at runtime. In this configuration, the
TensorFlow-based SLING parser runs at 200 tokens per CPU second.

To speed up parsing, we have created \emph{Myelin}, a
just-in-time compiler for neural networks that compiles network cells into
x64 machine code at runtime. The generated code exploits such
specialized CPU features as SSE, AVX, and FMA3, if available.
Tensor shapes and model parameters are fixed at runtime.
This allows us to optimize the network by folding constants, unrolling
loops, and pre-computing embeddings, among other transformations. The JIT compiler can also fix the data
instance layout at compile-time to speed up runtime data access.

The Myelin-based SLING parser runs at 2500 tokens per CPU second, more
than ten times faster than the TensorFlow-based version
(Table~\ref{tab:runtime}).

\begin{table}[!t]
\centering
\begin{tabular}{|l|r|r|r|}
\hline
Runtime    & Speed     & Runtime      & Load      \\
           &           & size         & time      \\
\hline
TF         &  200 TPS  & 37.000 KB    & 10 secs   \\
Myelin     & 2500 TPS  &    500 KB    & 0.5 secs  \\
\hline
\end{tabular}
\caption{Comparison between TensorFlow-based SLING parser runtime and
Myelin-based parser runtime using JIT compilation.
Speed is measured as tokens parsed per CPU second, i.e. user+sys in time(1).}
\label{tab:runtime}
\end{table}

\begin{figure}[t]
  \centering
  \includegraphics[width = 200pt]{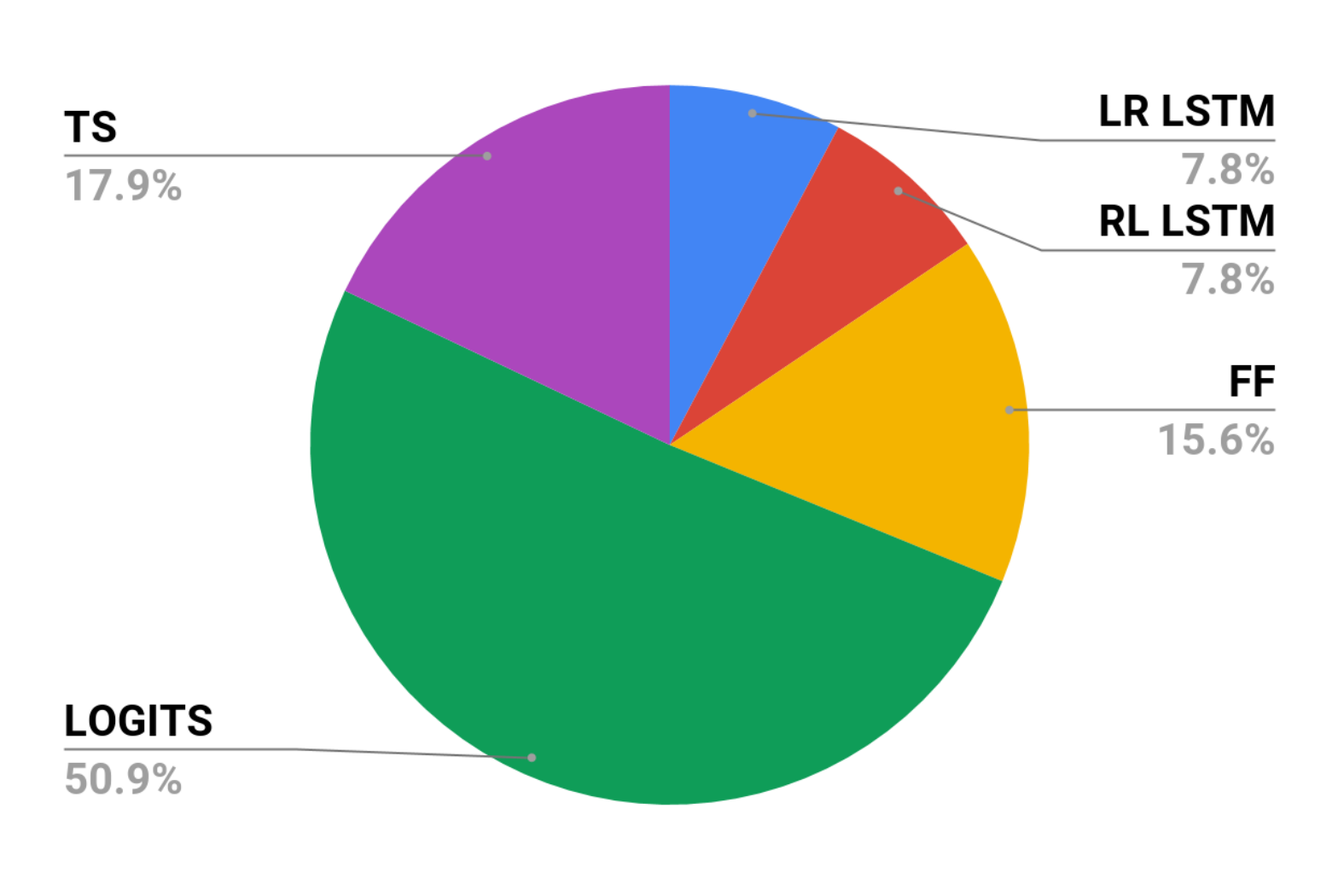}
  \caption{Runtime profile for running the Myelin-based SLING parser with the
  left-to-right LSTM (LR LSTM), right-to-left LSTM (RL LSTM), Feed-forward
  excluding logits (FF), Logits for output actions (LOGITS), and transition
  system and feature extraction (TS).}
  \label{fig:runtime}
\end{figure}

The Myelin-based SLING parser is independent of TensorFlow so it only needs to
link with the Myelin runtime (less than 500 KB) instead of the TensorFlow
runtime library (37 MB), and it is also much faster to initialize (0.5 seconds
including compilation time) than the TensorFlow-based parser (10 seconds).
Figure~\ref{fig:runtime} shows a breakdown of the CPU time for the Myelin-based
parser runtime.

Half the time is spent computing the logits for the output
actions. This is expensive because the OntoNotes-based corpus has 6968 actions,
where the vast majority of the actions are of a form like
{\bf EVOKE(/pb/hit-01, 1)}, one for each PropBank roleset predicate in the
training data. Table~\ref{tab:action-table} shows that only about 26\% of all the
actions are EVOKE actions. The output layer of the FF unit could be turned into
a cascaded classifier, where if the first classifier predicts a generic
{\bf EVOKE(/pb/predicate, 1)} action, it would use a secondary classifier to
predict the predicate type. This could almost double the speed of the parser.

\section{Conclusion}
\label{sec:conclusion}

We have described SLING, a framework for parsing natural language into
semantic frames. Our experiments show that it is feasible to build a
semantic parser that outputs frame graphs directly without any intervening
symbolic representation, only using the tokens as inputs.
We illustrated this on the joint task of predicting entity mentions, entity types,
measures, and semantic role labeling.
While the LSTMs and TBRUs are expensive to compute, we can achieve acceptable
parsing speed using the Myelin JIT compiler.
We hope to make use of SLING in the future for further exploration into
semantic parsing.

\section*{Acknowledgements}
\label{sec:ack}

We would like to thank Google for supporting us in this project and allowing us
to make SLING available to the public community. We would also like to thank the
Tensorflow and DRAGNN teams for making their systems publicly available.
Without it, we could not have made SLING open source.

\bibliography{sling}
\bibliographystyle{acl_natbib}

\end{document}